*Б.М. Павлишенко*

*Львівський національний університет імені Івана Франка*
*вул. Драгоманова, 50, 79005 Львів, Україна*
*e-mail:pavlsh@yahoo.com*

# МОДЕЛЬ РЕШІТКИ СЕМАНТИЧНИХ КОНЦЕПТІВ ДЛЯ ІНТЕЛЕКТУАЛЬНОГО АНАЛІЗУ МІКРОБЛОГІВ

В роботі запропонована модель решітки семантичних концептів для інтелектуального аналізу повідомлень мікроблогів. Показано, що використання цієї моделі є ефективним при аналізі семантичних зв'язків та виявленні асоціативних правил для ключових термінів.
Ключові слова: інтелектуальний аналіз даних, аналіз формальних концептів, мікроблоги, семантичні поля.

## Постановка проблеми

Методи сучасного інтелектуального аналізу даних ефективно використовуються в обробці контенту веб-ресурсів. Система мікроблогів Twitter є одним із популярних засобів взаємодії користувачів за допомогою коротких повідомлень (не більше 140 символів). Формат таких повідомлень є надзвичайно простий і дозволяє згадувати в тексті інших користувачів (наприклад, @username) та тематичні групи за допомогою хештегів з позначкою # (наприклад, #software). Повідомлення одночасно надсилаються згаданим в них користувачам та тематичним групам. Такий формат дає можливість за деяким ключовим словом виявляти повідомлення, які включають в себе це слово, а також виявляти користувачів та групи, які мають відношення до тематики заданої цим ключовим словом. Такі повідомлення також несуть інформацію про взаємозв'язок між окремими користувачами та ключовими словами. Для Twitter-повідомлень характерна висока густина тематично значимих ключових слів. Ця особливість зумовлює перспективність досліджень мікроблогів засобами інтелектуального аналізу та актуальність розвитку методів інтелектуального аналізу текстових повідомлень для виявлення семантичних зв'язків між основними поняттями та тематиками обговорень в мікроблогах.

## Аналіз останніх досліджень та публікацій

Інтелектуальний аналіз слабо структурованих даних, наприклад, текстових масивів є однією із складових частин сучасних інформаційних технологій [1,2]. В такому аналізі використовують, зокрема, алгоритми пошуку частин множин ознак та асоціативних правил, за допомогою яких можна виявити взаємозв`язок між підмножинами даних [3,4,5,6]. Одним із ефективних методів аналізу даних є теорія аналізу формальних концептів [2,3,4,5]. В цій теорії розглядають відношення об'єктів та їх атрибутів, на основі якого будують алгебраїчну решітку формальних концептів. Кожен концепт об'єднує множину об'єктів та їх спільних атрибутів. На основі частин множин спільних атрибутів виявляють асоціативні правила, які відображають зв'язки між атрибутами на множині аналізованих об'єктів. В роботі [8] використовують теорію аналізу формальних концептів для аналізу американських політичних блогів. Актуальним на даний час є створення моделі формальних концептів для аналізу мікроблогів, яка б враховувала семантичну структуру повідомлень. Для цього доцільно ввести поняття семантичного поля, яке б об'єднувало ключові лексеми тематики аналізу.

## Цілі статті

Створимо теоретико-множинну модель повідомлень мікроблогів. Розглянемо можливість дослідження повідомлень мікроблогів на основі методів аналізу формальних концептів, який базується на теорії алгебраїчних решіток [2,3,4,5]. Розглянемо утворення семантичних концептів та асоціативних правил. На основі утвореної моделі проаналізуємо тестовий масив повідомлень системи Twitter.

## Виклад основного матеріалу. Теоретична модель.

Розглянемо модель, яка описує повідомлення мікроблогів, їх словник, користувачів та тематичні групи. Нехай вибрано деяке ключове слово *kw*, яке задає тематику повідомлень і є наявне у всіх повідомленнях, наприклад *kw='software'*. Визначимо множину повідомлень мікроблогів:

$$TW^{kw} = \left\{ tw^{(kw)}{}_i \mid kw \in tw_i \right\}. \qquad (1)$$

Загальний словник аналізованого масиву повідомлень розглянемо як мультимножину

$$W_S^{tw(kw)} = \left\{ n_i^{st}(w_i) \mid w_i \in TW^{kw} \right\} \qquad (2)$$

де $n_i^{st}$ - кількість зустрічань лексеми $w_i$ в повідомленнях аналізованого масиву. Оскільки всі повідомлення містять наперед задане ключове слово (в наших дослідженнях це слово – «*software*»), то такий масив повідомлень буде охоплювати деякий наперед заданий семантичний спектр інформації. Вввведемо узагальнене поняття семантичного поля [7]. Під семантичним полем будемо розуміти деяку підмножину словника, елементи якої об'єднані деяким спільним семантичним поняттям. В загальному випадку такі поняття можуть об'єднувати ключові слова, які відносяться до підрозділів аналізованої тематики. Уведемо множину семантичного поля, в яку входять ключові слова, та хеш-теги назв тематичних груп

$$Keywords = \left\{ keyword_i \right\} . \qquad (3)$$

Множина *Keywords*, яка відображає задану тематику може бути сформована на основі експертного аналізу, коли експерт формує масив ключових слів, які охоплюють напрям досліджень. Семантичне поле може бути також утворене на основі знайдених частин множин лексем. Такі множини формуються із наборів лексем, які одночасно зустрічаються у повідомленнях із частотою, більшою за деякий заданий поріг. Очевидно, що деяка підмножина масиву частин множин ключових лексем буде відображати семантику напрямку досліджень мікроблогів.

Використовуючи теорію аналізу формальних концептів [2,3,4,5] розглянемо формальний контекст як трійку

$$K^{tw(kw)} = \left(TW_s^{(kw)}, Keywords, I_s \right) \tag{4}$$

де $I_s$ - відношення $I_s \subseteq TW_s^{(kw)} \times Keywords$, яке описує зв'язки повідомлень із ключовими лексемами в цих повідомленнях. Вважаємо, що $(tw_i^{(kw)}, keyword_j) \in I_s$, якщо термін $keyword_j$ зустрічається в повідомленні $tw_i^{(kw)}$. Відношення $I_s$ можна розглядати як множину

$$I_s = \left\{ (tw_i, keyword_j) \mid keyword_j \in tw_i^{(kw)} \right\}. \tag{5}$$

Уведемо решітку семантичних концептів. Для деяких $Ext \subseteq TW_s^{(kw)}$, $Int \subseteq Keywords$ визначимо такі відображення

$$Ext' = \left\{ keyword_j \in Keywords \mid tw_i^{(kw)} \in Ext : (tw_i^{(kw)}, keyword_j) \in I_s \right\} \tag{6}$$

$$Int' = \left\{ tw_i^{(kw)} \in TW_s^{(kw)} \mid keyword_j \in Int : (tw_i^{(kw)}, keyword_j) \in I_s \right\} \tag{7}$$

Множина $Ext'$ описує ключові терміни, які властиві документам множини $Ext$, а множина $Int'$ описує повідомлення, які містять ключові терміни множини $Int$. Уведемо семантичний концепт як пару

$$Concept = (Ext, Int), \tag{8}$$

до якої належать повідомлення з множини $Ext \subseteq TW_s^{(kw)}$ та ключові терміни з множини $Int \subseteq Keywords$ з такими умовами

$$\begin{cases} Ext' = Int, \\ Int' = Ext. \end{cases} \tag{9}$$

Множину $Ext$ назвемо об'ємом, а $Int-$ змістом семантичного концепту $Concept$. В семантичному контексті $K^{tw(kw)}$ утворюється частково-впорядкована множина семантичних концептів

$$\Psi(TW_s^{(kw)}, Keywords, I_s) = \{ Concept_m = (Ext_m, Int_m) \}, \quad (10)$$

Семантичний концепт

$$Concept_1 = (Ext_1, Int_1) \quad (11)$$

є менш загальним за об'ємом чим концепт

$$Concept_2 = (Ext_2, Int_2) \quad (12)$$

тобто виконується умова

$$(Ext_1, Int_1) \leq (Ext_2, Int_2), \quad (13)$$

якщо

$$Ext_1 \subseteq Ext_2 \Leftrightarrow Int_1 \supseteq Int_2. \quad (14)$$

В цьому випадку концепт $Concept_2$ можна вважати узагальненням концепту $Concept_1$. Семантичний концепт можна розглядати як підматрицю семантичного контексту, яка повністю заповнена одиницями. Решітку концептів часто відображають за допомогою діаграм Гассе. В аналізі семантичного контексту $K^{tw(kw)}$ кожний елемент діаграми представляє семантичний концепт. Такі діаграми відображають внутрішню семантичну структурну організацію повідомлень користувачів та відповідних їм груп ключових термів.

Розглянемо поняття порядкового ідеалу та фільтра для деякої частково впорядкованої множини $(P, \leq)$. Порядковим ідеалом називають підмножину $J \subseteq P$ для якої

$$\forall x \in J, y \leq x \Rightarrow y \in J. \quad (15)$$

Порядковим фільтром називають підмножину $F \subseteq P$ для якої

$$\forall x \in F, y \geq x \Rightarrow y \in F. \quad (16)$$

Використання понять порядкового ідеала та фільтра може бути ефективним в аналізі решітки семантичних концептів. Порядковим ідеалом деякого концепта будуть концепти, які пов'язані з ним на діаграмі Гассе і знаходяться нижче нього включаючи концепт, який відповідає інфімуму решітки. Порядковим фільтром деякого концепту є множина пов'язаних із ним концептів, які знаходяться вище нього в решітці, включаючи концепт, який відповідає супремуму решітки. Зміст деякого концепту є підмножиною змістів концептів, які належать до його порядкового ідеалу. З іншої сторони, об'єднання змістів концептів, які утворюють порядковий фільтр деякого концепту утворює зміст цього концепту. Інформативним для аналізу є також розгляд об'єднання порядкового фільтра та ідеала. Множина змістів такого об'єднання утворює деяке семантичне поле, яке відображає множину взаємопов'язаних понять. В одній решітці може знаходитись декілька таких незалежних об'єднань порядкових ідеалів та фільтрів. Отже, одним із методів формування семантичних полів є пошук множини змістів концептів деякого об'єднання ідеала та фільтра заданого формального контексту.

На основі розрахованої решітки семантичних концептів можна виявити асоціативні правила, які відображають семантичні структурні зв'язки між ключовими словами. Під асоціативним правилом деякого контексту $K^{tw(kw)} = \left(TW_s^{(kw)}, Keywords, I_s\right)$ будемо розуміти вираз

$$A \rightarrow B, \quad A, B \subseteq Keywords \tag{17}$$

Підмножину $A$ називають передумовою, а $B$ – наслідком асоціативного правила $A \rightarrow B$. Важливими характеристиками асоціативних правил є підтримка (support) $Supp_{A \rightarrow B}$ та достовірність (confidence) $Conf_{A \rightarrow B}$, які можна обрахувати за такими виразами:

$$Supp_{A \rightarrow B} = \frac{|(A \cup B)'|}{|TW_s^{(kw)}|} \tag{18}$$

$$Conf_{A \rightarrow B} = \frac{|(A \cup B)'|}{|A'|} \tag{19}$$

У випадку коли $Conf_{A \rightarrow B} = 1$ асоціативне правило (17) є імплікацією, тобто виконується завжди, коли зустрічається передумова $A$. Значення $Supp_{A \rightarrow B}$ характеризує частку повідомлень $TW_s^{(kw)}$, яка містить ознаки $A \cup B$. Величина $Conf_{A \rightarrow B}$ характеризує частку повідомлень із ключовими словами множини $A$, яка також містить ключові слова множини $B$. Актуальними для аналізу є правила із деяким заданим мінімальним значенням достовірності та підтримки:

$$Supp_{A \rightarrow B} > Supp_{\min} \tag{20}$$
$$Conf_{A \rightarrow B} > Conf_{\min} \tag{21}$$

Асоціативні правила із умовами (20)-(21) називають частими та отримують із частої підмножини ключових лексем:

$$F \subseteq Keywords, \quad |F'| > \theta, \quad F = A \cup B, \quad A \cap B = \varnothing, \tag{22}$$

де $\theta$ - деякий поріг частої множини.

## Експериментальна частина

Для реалізації експериментальних досліджень розроблено пакет прикладних програм на мові Perl. За допомогою цього пакету використовуючи API системи Twitter завантажено тестовий масив повідомлень, які містять ключове слово "software" а також хеш-тег "#software". Тобто, відібрано повідомлення заданого тематичного напряму пов'язаного із програмним забезпеченням. Твіти з ключовим словом 'software' завантажувались в період з 06.08.11 по 11.08.11. В загальному завантажено 75977 твітів. Далі проведена фільтрація твітів і взято до розгляду лише лексеми, які повторюються не менше 10 раз і не більше 4000 раз. Наведемо приклади високочастотних лексем в порядку спадання частоти зустрічань

#software (3371), engineer (3186), download (2615), #jobs (2279), online (2098), business (1865), marketing (1758), windows (1751), development (1704), developer (1673), management (1525)

Отриманий частотний словник містить 6325 лексем. Були відфільтровані високочастотні стоп-слова, які не несуть семантичну інформацію. Знайдені часті множини термінів із підтримкою більше 10. До розгляду були взяті твіти, які містили не менше 5 лексем. Також розглядались часті множини із кількістю термінів від 2 до 5. Отримано список із 2879 частих множин, які відповідають наведеним вище умовам. Для зменшення кількості частих множин було збільшено мінімальну підтримку частих множин до 20. В цьому випадку кількість частих множин зменшилась до 1049. Наведемо деякі із них

{manager, #job}, {computer, windows}, {#job, developer}, {microsoft, windows}, {security, internet}, {looking, #job}, {player, traffic, video}, {script, #php}, {servers, hosting}, {servers, remote, desktop, hosting}, {salary, #hiring, location, #job}, {blackberry, android}

В аналізі розглядались решітки формальних концептів для семантичних полів різного розміру та змісту. Розглянемо твіти в яких присутні лексеми такого найпростішого семантичного поля $S_1$

{london, lndia, windows, microsoft, android, #mysql, scripts, #linux, #job, developer }

В це семантичне поле включено географічні назви, операційні системи, хеш-тег #job . Решітка семантичних концептів буде відображати взаємозв'язок цих понять в повідомленнях мікроблогів. Після фільтрації масиву вхідних повідомлень за наведеним семантичним полем отримано масив із 8920 твітів. Для розрахунку решітки концептів та побудови діаграм Гассе був використаний пакет программ Lattice Miner. На рис.1 наведено діаграму Гассе, яка відображає утворену решітку семантичних концептів для семантичного поля $S_1$. На цій діаграмі наведено зміст концептів верхнього рівня. Змісти концептів нижніх рівнів є комбінаціями наведених змістів відповідно до зв'язків на діаграмі.

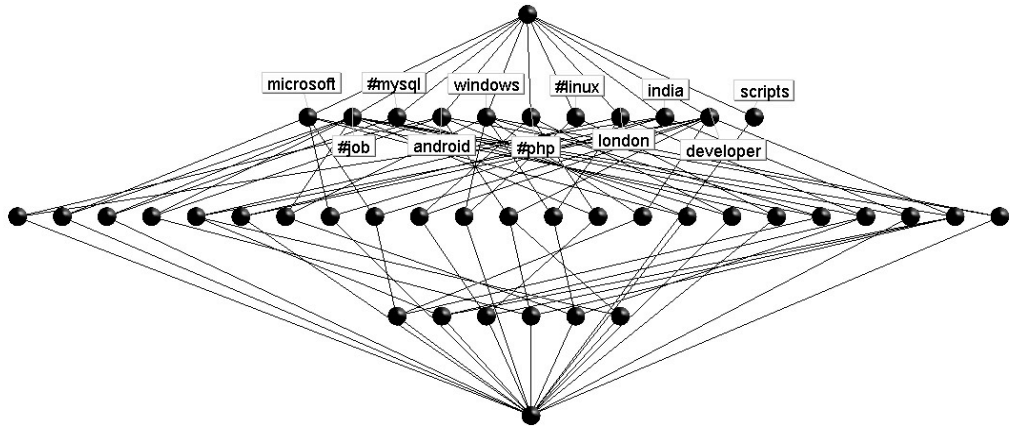

*Рис.1 Діаграма Гассе для решітки семантичних концептів семантичного поля $S_1$.*

На рис.2 виділено фільтр та ідеал для концепта {android, developer}. Для концептів наведено зміст та об'єм в процентах. Концепт інфімуму не наведено, оскільки він містить нульовий об'єм.

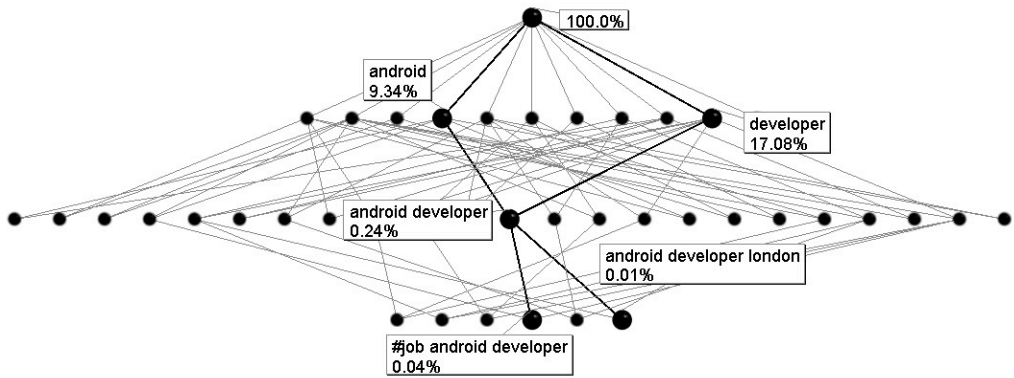

*Рис.2 Порядковий фільтр та ідеал для концепта {android, developer}.*

Таблиця 1 містить приклади асоціативних правил та їх кількісні характеристики для наведеної на рис.1 решітки семантичних концептів.

*Таблиця 1.*

**Асоціативні правила відфільтрованого за семантичним полем $S_1$ масиву повідомлень.**

| № | Передумова $A$ | Наслідок $B$ | $Supp_{A \rightarrow B}$ | $Conf_{A \rightarrow B}$ |
|---|---|---|---|---|
| 1 | {#php} | {#mysql} | 0.33% | 22.55% |
| 2 | {london} | {#job} | 0.58% | 20.47% |
| 3 | {#job, android} | {developer} | 0.04% | 50.0% |

| 4 | {#mysql, #php} | {script} | 0.2% | 60.0% |
| 5 | {linux, mysql} | {developer} | 0.04% | 30.76% |
| 6 | {android, london} | {developer} | 0.01% | 100.0% |

Розглянемо інше більш об'ємніше семантичне поле $S_2$, яке складається із таких лексем та хеш-тегів:

{#linux, #opensource, android, browser, center, drivers, earth, features, google, greater, installs, internet, iphone, latest, leader, linux, netscape, phones, popular, powerful, printer, sales, telemarketing, tracking, ubuntu}

Для цього семантичного поля отримано відфільтрований контекст, який містить 4681 твітів. Розрахована решітка семантичних концептів наведена на рис. 3.

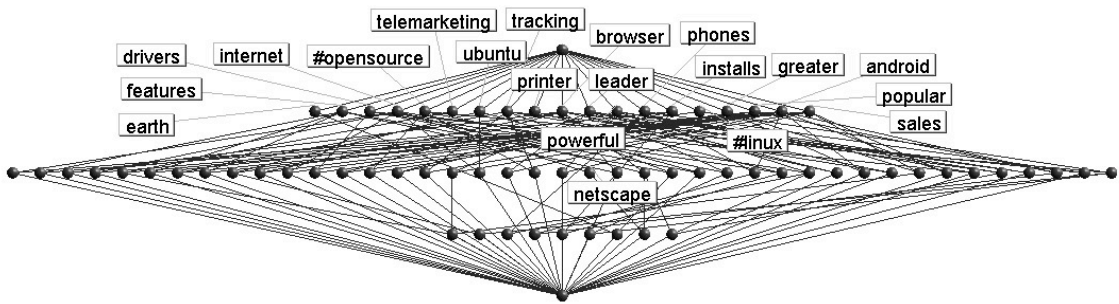

*Рис.3 Діаграма Гассе для решітки семантичних концептів семантичного поля $S_2$.*

На рис.4 показано зв'язки для концепта {android}, які відображають його порядковий фільтр та ідеал. Фільтр представлений лише концептом супремуму та самим концептом {android}.

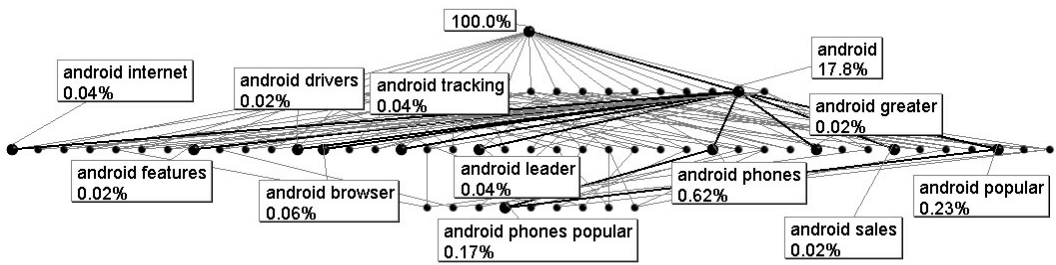

*Рис.4 Порядковий фільтр та ідеал для концепта {android}.*

Для наведеного вище семантичного поля $S_2$ розраховано асоціативні правила, приклади яких наведено в таблиці 2.

*Таблиця 2.*

**Асоціативні правила відфільтрованого
за семантичним полем $S_2$ массиву повідомлень.**

| № | Передумова A | Наслідок B | $Supp_{A \to B}$ | $Conf_{A \to B}$ |
|---|---|---|---|---|
| 1 | {#linux} | {#opensource} | 1.62% | 45.5% |
| 2 | {telemarketing} | {sales} | 0.21% | 83.33% |
| 3 | {browser, internet} | {netscape} | 0.21% | 55.55% |
| 4 | {#linux, ubuntu} | {#opensource} | 0.14% | 46.66% |
| 5 | {android, popular} | {phones} | 0.17% | 72.72% |
| 6 | {#opensource, ubuntu} | {#linux} | 0.14% | 100.0% |
| 7 | {phones, popular} | {android} | 0.17% | 100.0% |

Серед наведених в таблиці 2 правил можна виявити такі імплікації:

$$\{\#opensource, ubuntu\} \Rightarrow \{\#linux\}; \quad \{phones, popular\} \Rightarrow \{android\} \qquad (23)$$

Правила (23) є імплікаціями лише для відфільтрованого масиву твітів і в загальному випадку повідомлень мікроблогів можуть не бути імплікаціями.

## Висновки

Застосування теорії аналізу формальних концептів є ефективним в інтелектуальній обробці повідомлень мікроблогів. Використання моделі решітки семантичних концептів дає можливість аналізувати семантично зв'язані множини лексем та будувати асоціативні правила. Формування семантичних полів на основі масиву виявлених частин множин дає можливість суттєво звузити пошук асоціативних правил та розмір решітки семантичних концептів в алгоритмах інтелектуального аналізу текстів.

## Література

**THE MODEL OF SEMANTIC CONCEPTS LATTICE FOR
DATA MINING OF MICROBLOGS**

*B.M. Pavlyshenko*

*Ivan Franko Lviv National University,
Dragomanov Str. 50, Lviv, UA–79005 Ukraine, e-mail:pavlsh@yahoo.com*

The model of semantic concept lattice for data mining of microblogs has been proposed in this work. It is shown that the use of this model is effective for the semantic relations analysis and for the detection of associative rules of key words.
Key words: data mining, formal concepts analysis, FCA, microblogs, semantic fields.

**МОДЕЛЬ РЕШІТКИ СЕМАНТИЧНИХ КОНЦЕПТІВ ДЛЯ
ІНТЕЛЕКТУАЛЬНОГО АНАЛІЗУ МІКРОБЛОГІВ**

*Б.М. Павлишенко*

*Львівський національний університет імені Івана Франка
вул. Драгоманова, 50, 79005 Львів, Україна, e-mail:pavlsh@yahoo.com*

В роботі запропонована модель решітки семантичних концептів для інтелектуального аналізу повідомлень мікроблогів. Показано, що використання цієї моделі є ефективним при аналізі семантичних зв'язків та виявленні асоціативних правил для ключових термінів.
Ключові слова: інтелектуальний аналіз даних, аналіз формальних концептів, мікроблоги, семантичні поля.

**МОДЕЛЬ РЕШЕТКИ СЕМАНТИЧЕСКИХ КОНЦЕПТОВ ДЛЯ
ИНТЕЛЛЕКТУАЛЬНОГО АНАЛИЗА МИКРОБЛОГОВ**

*Б.М. Павлышенко*

*Львовский национальный университет имени Ивана Франко
ул. Драгоманова, 50, 79005 Львов, Украина, e-mail:pavlsh@yahoo.com*

В работе предложена модель решетки семантических концептов для интеллектуального анализа сообщений микроблогов. Показано, что использование этой модели эффективно при анализе семантических связей и выявлении ассоциативных правил для ключевых терминов.
Ключевые слова: интеллектуальный анализ данных, анализ формальных концептов, микроблоги, семантические поля.

*B.M. Pavlyshenko*
*Ivan Franko Lviv National University,*
*Dragomanov Str. 50, Lviv, UA–79005 Ukraine, e-mail:pavlsh@yahoo.com*

The methods of modern data mining are used effectively in Web content resources processing. The system of microblogs Twitter is one of the most popular for users' interaction with the help of short messages. The model of semantic concept lattice for data mining of microblogs has been proposed in this work. It is shown that the use of this model is effective for the semantic relations analysis and for the detection of associative rules of keywords in the microblogs messages array. For the experimental research the package of applied programs in the language Perl has been developed. With the help of this package and using the API of Twitter the test array of messages that contain the word "software" and the hash tag "# software" has been downloaded. A set of thematic messages associated with the software themes has been selected. The lattice of formal concepts for the semantic fields of different size and content has been considered. The tweets containing lexemes of different semantic fields have been analysed. The semantic concepts lattice reflect the interaction of concepts in microblogs messages. After filtering the array of input messages by given semantic field, there was received an array of 8920 tweets. The package of programs Lattice Miner was used for calculating the concepts lattice. On the basis of concepts lattice the associative rules that represent the relations between semantic concepts of analysed subjects have been found. The application of the theory of formal concept analysis is effective in the processing of intellectual microblogs messages. The use of lattice models of semantic concepts allows to analyse the sets of lexemes, that are semantically related, and to construct associative rules. The formation of semantic fields based on the array of identified frequent sets enables to narrow significantly the search of associative rules and lattice size of semantic concepts in algorithms of text mining.

## МОДЕЛЬ РЕШЕТКИ СЕМАНТИЧЕСКИХ КОНЦЕПТОВ ДЛЯ ИНТЕЛЛЕКТУАЛЬНОГО АНАЛИЗА МИКРОБЛОГОВ

**Б.М. Павлышенко**
*Львовский национальный университет имени Ивана Франко*
*ул. Драгоманова, 50, 79005 Львов, Украина, e-mail:pavlsh@yahoo.com*

Методы современного интеллектуального анализа данных эффективно используются в обработке контента веб-ресурсов. Система микроблогов Twitter является одним из популярных средств взаимодействия пользователей с помощью коротких сообщений. В работе предложена модель решетки семантических концептов для интеллектуального анализа сообщений микроблогов. Показано, что использование этой модели эффективно при анализе семантических связей и выявлении ассоциативных правил для ключевых терминов в массиве сообщений микроблогов. Для реализации экспериментальных исследований разработан пакет прикладных программ на языке Perl. С помощью этого пакета используя API системы Twitter скачано тестовый массив сообщений, содержащих ключевое слово "software", а также хеш-тег "# software". Отобраны сообщения заданного тематического направления связанного с программным обеспечением. В анализе рассматривались решетки формальных концептов для семантических полей разного размера и содержания. Рассмотрены твиты, в которых присутствуют лексемы разных семантических полей. Решетка семантических концептов отображает взаимосвязь понятий в сообщениях микроблогов. После фильтрации массива входящих сообщений этим семантическим полем, получено массив из 8920 твитов. Для расчета решетки был использован пакет программ Lattice Miner. На основе полученной решетки концептов обнаружено ассоциативные правила, отражающие связи между семантическими понятиями рассматриваемой тематики. Применение теории анализа формальных концептов является эффективным в интеллектуальной обработке сообщений микроблогов. Использование модели решетки семантических концептов позволяет анализировать семантически связанные множества лексем и строить ассоциативные правила. Формирование семантических полей на основе массива выявленных частых множеств дает возможность существенно сузить поиск ассоциативных правил и размер решетки семантических концептов в алгоритмах интеллектуального анализа текстов.

**Відомості про автора:**
Павлишенко Богдан Михайлович
кандидат фізико-математичних наук,
доцент факультету електроніки Львівського національного університету імені Івана Франка.
Службова адреса: факультет електроніки, вул. Драгоманова, 50, Львів, 79005, Україна.
моб.т.: 0505037290, e-mail: pavlsh@yahoo.com